\pdfoutput=1

\PassOptionsToPackage{table,dvipsnames}{xcolor}

\documentclass[11pt]{article}

\usepackage[final]{acl}

\usepackage{times}
\usepackage{latexsym}
\usepackage{xcolor}  
\usepackage[T1]{fontenc}

\usepackage[utf8]{inputenc}

\usepackage{microtype}

\usepackage{inconsolata}

\usepackage{graphicx}

\title{LCDS: A Logic-Controlled Discharge Summary Generation System Supporting Source Attribution and Expert Review}

\author{
    Cheng Yuan$^{1}$\thanks{Equal Contribution.}, 
    Xinkai Rui$^{1,2*}$, 
    \textbf{Yongqi Fan}$^{1}$,
    \textbf{Yawei Fan}$^{1}$,
    \textbf{Boyang Zhong}$^{1}$, \\
    \textbf{Jiacheng Wang}$^{1}$, 
    \textbf{Weiyan Zhang}$^{1}$\thanks{Co-corresponding Author.}, 
    \textbf{Tong Ruan}$^{1\dagger}$ \\
    $^1$East China University of Science and Technology, Shanghai 200237, China \\ 
    $^2$Ruijin Hospital, Shanghai Jiaotong University School of Medicine, \\ Shanghai 200025, China \\
    {\tt \{ruantong,weiyanzhang\}@ecust.edu.cn} \\
}

\begin{document}
\maketitle


\begin{abstract}
Despite the remarkable performance of Large Language Models (LLMs) in automated discharge summary generation, they still suffer from hallucination issues, such as generating inaccurate content or fabricating information without valid sources. In addition, electronic medical records (EMRs) typically consist of long-form data, making it challenging for LLMs to attribute the generated content to the sources. To address these challenges, we propose \textbf{LCDS}, a \textbf{L}ogic-\textbf{C}ontrolled \textbf{D}ischarge \textbf{S}ummary generation system. LCDS constructs a source mapping table by calculating textual similarity between EMRs and discharge summaries to constrain the scope of summarized content. Moreover, LCDS incorporates a comprehensive set of logical rules, enabling it to generate more reliable silver discharge summaries tailored to different clinical fields. Furthermore, LCDS supports source attribution for generated content, allowing experts to efficiently review, provide feedback, and rectify errors. The resulting golden discharge summaries are subsequently recorded for incremental fine-tuning of LLMs.
Our project and demo video are in the GitHub repository \url{https://github.com/ycycyc02/LCDS}.
\end{abstract}

\section{Introduction}
The discharge summary (DS) is the final section of an electronic medical record (EMR) that consolidates essential patient information, such as admission details, medical history, diagnoses, treatments, medications, and follow-up recommendations~\cite{xiong2019study}. It plays a critical role in ensuring continuity of patient care, facilitating communication between healthcare providers and patients, and supporting clinical decisions~\cite{lenert2014rethinking, kripalani2007promoting, li2013timeliness, vanwalraven2002effect}. Traditionally, discharge summaries are manually written by physicians, making the process time-consuming, labor-intensive, and susceptible to subjective biases~\cite{xu2024overview, hartman2023method, rink2023stressors}. Recently, large language models (LLMs) have shown great promise in automating discharge summary generation by leveraging retrieval, reasoning, and fine-tuning techniques~\cite{van2024adapted}. For example, \citet{liu2022retrieve} propose Re3Writer, which simulates physician workflows through medical knowledge retrieval and reasoning. Similarly, \citet{lyu2024uf} integrate extractive methods with generative techniques, combining named entity recognition (NER) and prompt-tuned text generation.

Despite these advancements, several critical challenges remain in automated discharge summary generation using LLMs.

\textbf{Precise Content Localization:} EMRs typically consist of long-form, complex, and heterogeneous data spanning multiple sections~\cite{wu2024epfl}. Directly feeding complete EMRs into LLMs can exceed their context limits, thus degrading the quality of generated summaries and increasing interference from irrelevant or redundant information.

\textbf{Accuracy and Hallucination Reduce:} Although LLMs demonstrate remarkable performance, they still suffer from hallucination issues, generating inaccurate or fabricated content lacking valid sources~\cite{maynez2020faithfulness, zhang2023famesumm, ji2023survey}. In the medical domain, this can significantly compromise patient safety and care quality. Effective strategies to impose logical constraints to mitigate these hallucinations remain underexplored.

\textbf{Adaptability to Different Clinical Departments:} While discharge summaries share a general structure across medical specialties, their detailed content requirements vary significantly. Current automated generation methods often lack adaptability to specific departmental needs, risking the omission of crucial clinical information.

\textbf{Traceability and Trustworthiness:} As discharge summaries directly influence patient care decisions, medication guidance, and follow-up treatments, ensuring content traceability is essential. However, current LLM-based generation systems lack explicit source attribution mechanisms, making it challenging for medical professionals to verify and trust the generated content.
To address these challenges, we propose A \textbf{LCDS} (\textbf{L}ogic-\textbf{C}ontrolled \textbf{D}ischarge \textbf{S}ummary Generation) System, featuring source attribution, logical constraints, and expert review:

\begin{itemize}
    \item \textbf{Source Mapping for Precise Content Localization}: LCDS constructs a source mapping table by calculating textual similarity between EMRs and discharge summaries, effectively constraining content selection and enhancing summary accuracy.

    \item \textbf{Logic-Controlled Summary Generation}: LCDS incorporates structured prompts guided by medical-domain logical rules, significantly improving factual accuracy and reducing hallucinations in generated discharge summaries.

    \item \textbf{Attribution-Based Expert Review}: LCDS segments generated summaries at the sentence level, explicitly attributing content to original EMR sources. This mechanism supports expert verification, facilitates error correction, and enhances clinical reliability.
\end{itemize}

Our system implements all proposed functionalities, demonstrating a complete pipeline for discharge summary generation from EMRs. Moreover, we conducted experiments using real-world clinical data from 15 medical departments. Experimental results show that LCDS outperforms existing methods in terms of accuracy, coherence, and clinical applicability of the generated discharge summaries, significantly reducing hallucinations and improving content traceability.




\section{Related Work}
Existing methods for automatic DS generation fall into three categories:

\textbf{Extraction-Abstracting Methods: }These methods first extract key information from medical records and then generate summaries, aiming to balance traceability and textual fluency. Representative studies include~\cite{shing2021towards, hartman2023automate, krishna2021generating}. While such approaches enhance factual accuracy, they heavily rely on the quality of the source text, making them prone to information omission.

\textbf{Knowledge-Enhanced Methods:} This category integrates external knowledge bases or retrieval-augmented techniques to improve the reliability of summaries. Examples include reinforcement learning-based medical entity verification \cite{zhang2020optimizing}, embedded entity retrieval alignment \cite{adams2024speer}, and a three-step generate framework comprising retrieval, reasoning, and synthesis \cite{liu2022retrieve}. However, these methods are computationally complex and constrained by the timeliness of the knowledge base.

\textbf{LLM-Based Methods:} These approaches leverage prompt engineering or fine-tuning techniques to adapt large models for medical applications. \cite{clough2024transforming} has shown that GPT-4 and its variants can generate summaries approaching physician-level quality. However, as noted by~\cite{williams2024evaluating, dubinski2024leveraging, kim2024patient}, the generated content still requires human review to ensure clinical accuracy. Additionally, LLMs are prone to hallucinations, potentially producing misleading or erroneous information. The lack of a clear provenance mechanism further complicates the verification of generated summaries by medical professionals.

\begin{figure*}[t]
  \includegraphics[width=\textwidth]{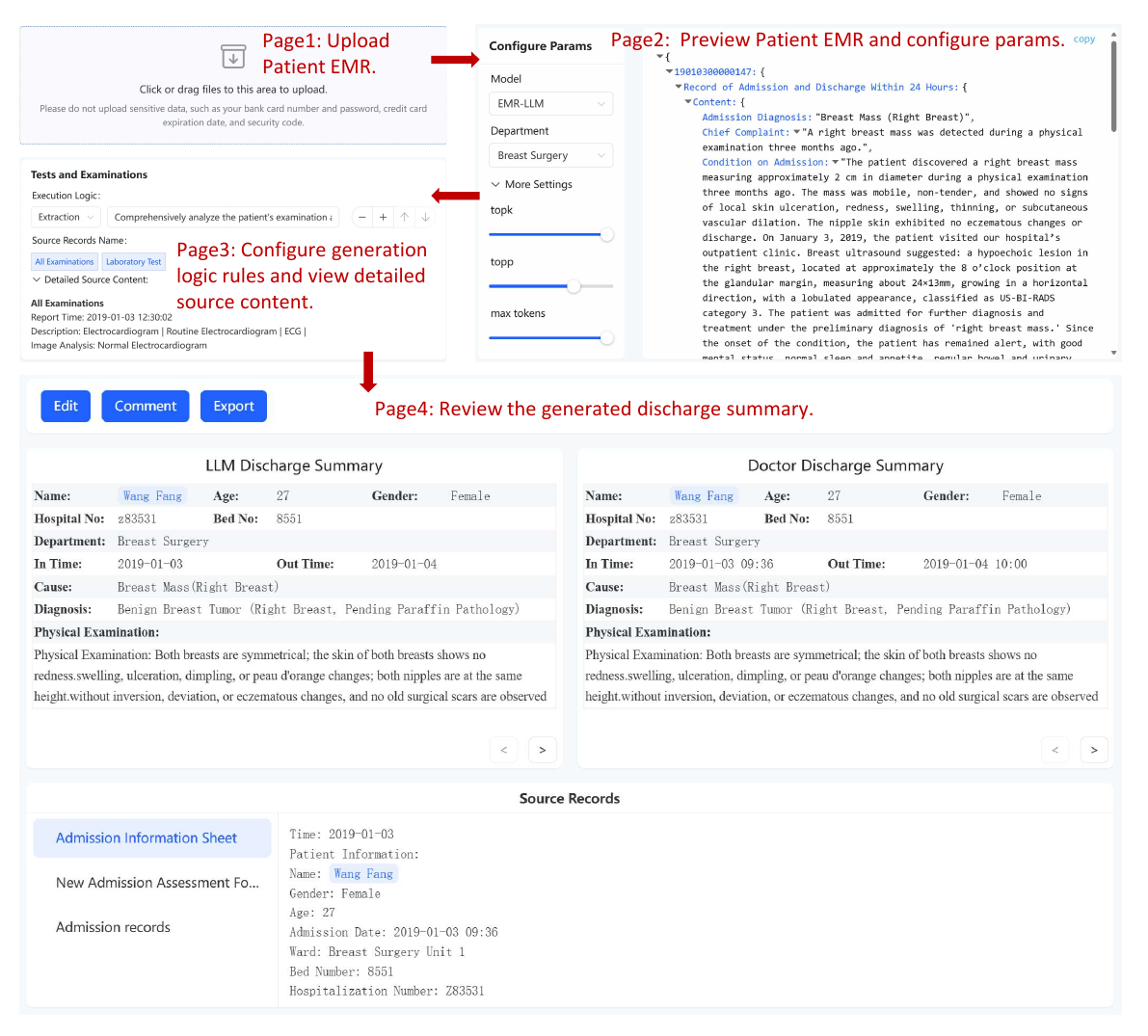}
  \caption{Screenshot of the LCDS web application, where the page functions are annotated.}
  \label{fig:pages}
\end{figure*}

\begin{figure*}[t]
  \includegraphics[width=\textwidth]{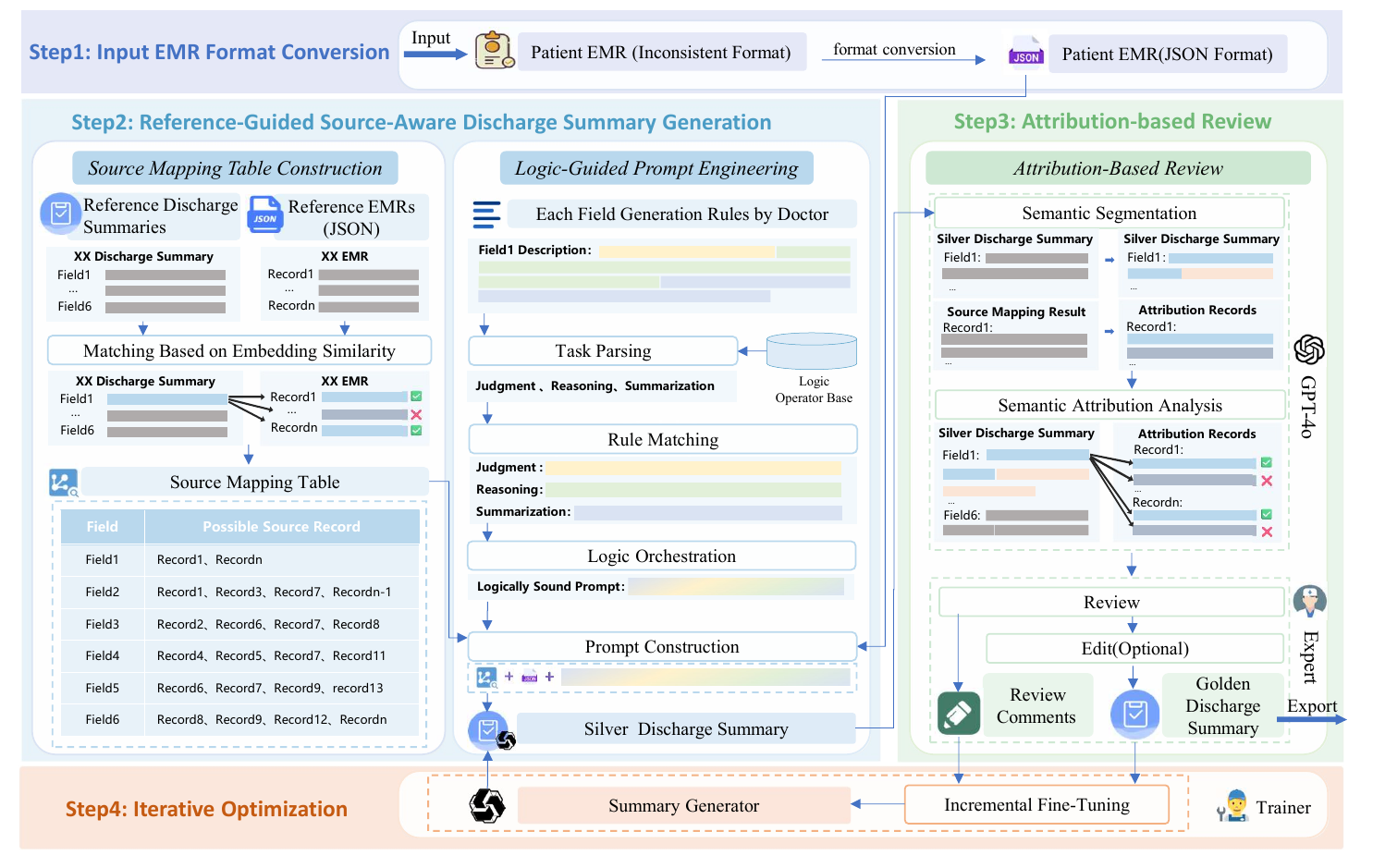}
  \caption{System workflow overview. The process includes four steps: (1) Upload and convert EMRs; (2) Extract key information, configure generation logic, and generate the discharge summary; (3) Perform attribution analysis and review; (4) Construct an incremental dataset and perform incremental learning.}
  \label{fig:framework}
\end{figure*}

\section{System Workflow and Usage Example}

This section introduces the system's usage and functionality through case studies. As shown in Figure \ref{fig:framework}, the workflow consists of four steps:

\textbf{Input EMR Format Conversion}: LCDS converts various types of EMR documents uploaded by users into a unified JSON format, ensuring data consistency and standardization.

\textbf{Reference-Guided Source-Aware Discharge Summary Generation}: Key content is extracted from standardized EMRs, and a “Silver” DS is generated based on refined logical field constraints.

\textbf{Attribution-Based Comparison and Review}: LCDS aligns each sentence in the summary with the original EMR, allowing experts to review, compare, and modify content for a high-quality ``Gold'' Discharge Summary.

\textbf{Iterative Optimization}: Review feedback and finalized discharge summaries create an incremental training dataset for continuous model optimization once enough data is accumulated.

\subsection{Input EMR Format Conversion}
As shown in Figure \ref{fig:pages}, users begin on Page 1 by uploading multiple EMR documents via a drag-and-drop interface (see Appendix~\ref{sec:appendixA} for supported document types). LCDS preprocesses and converts these documents into a unified JSON format, facilitating consistency and accurate source attribution. The unified format simplifies downstream processing and improves processing efficiency. Upon successful conversion, users proceed to Page 2, where the right panel displays structured EMR data, summarizing all uploaded records, and the left panel offers configuration options for model selection and department-specific logical rules, allowing users to tailor generation parameters to clinical needs.

\subsection{Reference-Guided Source-Aware Discharge Summary Generation}
After configuration, users proceed to Page 3, where they can preview source document names, extracted key content and customize logical constraints. LCDS supports 15 medical departments, with baseline source references provided for each DS field. As shown in Page 3 of Figure~\ref{fig:pages}, the \textit{“Source Records Name”} section displays source documents for the breast surgery department’s DS, while \textit{“Detailed Source Content”} shows extracted medical content. Users can modify logical rules in the \textit{“Execution Logic”} section, which supports extraction, reasoning, summarization, and judgment logic types. The fifth logic type, \textit{knowledge}, generates follow-up medication recommendations based on predefined mappings of medical history and test results to department-specific guidelines.

\subsection{Attribution-Based Comparison and Review}
After configuring Page 3, LCDS generates the ``Silver'' DS and redirects users to the comparison interface on Page 4. The upper section displays the generated summary on the left, with physician-authored summaries for comparison. The lower section lists the source documents and their contents. Users can hover over the generated summary to highlight the matching content in the physician-written summary. Clicking on any part updates the lower section to show the corresponding source document and highlights related sentences. The top toolbar provides Edit, Comment, and Export functions for experts to modify content, annotate feedback, and download the final ``Golden'' DS in JSON format.

\subsection{Iterative Optimization}
Through the aforementioned steps, LCDS accumulates a dataset of ``Silver'' DSs and expert-reviewed ``Golden'' counterparts, which serves as an incremental training corpus for continuous model refinement. As data accumulates, trainers use these revised summaries for ongoing model improvement.

\section{System Overview}

\subsection{Summary Generator}
In our work, we utilize ChatGLM3-6B~\cite{glm2024chatglm} to generate DSs. To enhance the model's understanding of task details and improve its performance in this text generation task, we construct a high-quality instruction dataset and fine-tune the model using LoRA.\footnote{We provide some examples of instruction dataset in \url{https://github.com/ycycyc02/LCDS}.} The fine-tuned model is named EMRLLM. Since our backend model is modular, we can also replace EMRLLM with other LLMs such as Alpacare~\cite{zhang2023alpacare}, Bentaso~\cite{wang2023huatuo}, or HuatuoGPT~\cite{zhang2023huatuogpt}.

\subsection{Source Mapping Table Construction}

To enhance input precision, minimize hallucinations caused by excessive text scope, and improve the efficiency and accuracy of information localization, we construct a DS-EMR mapping table, which clearly defines the relationships between the DS and its corresponding source documents and relevant fields.

We collect 500 EMRs from 15 departments, each containing a physician-authored DS. These DSs serve as ground truth for localizing information from the corresponding source documents.
To facilitate structured generation, we divide each DS into six distinct Fields: (1) Patient Information, (2) Discharge Diagnosis, (3) Tests and Examinations, (4) Disease Course and Treatment, (5) Condition at Discharge, and (6) Post-Discharge Medication Advice.

For short-text Fields such as ``patient information'', we directly use the ground truth as a keyword to search across all fields of the medical records. If a field contains the keyword, it is identified as the corresponding information source.

For long-text Fields such as ``Disease Course and Treatment'', content may originate from multiple medical records, and different sentences may correspond to different source documents. To address this, we first perform sentence-level semantic segmentation and then determine the source of each segment. Specifically, we employ in-context learning (ICL) for semantic segmentation, where the input consists of the ``Disease Course and Treatment'' text, and the output includes categorized labels and their corresponding content. For instance, if a patient’s disease course involves surgery, chemotherapy, pathology, and discharge details, the output should be \{Surgery: corresponding surgical description, Chemotherapy: corresponding chemotherapy description, Pathology: corresponding pathology description, Discharge Details: corresponding discharge description\}. Using this approach, we break down long texts into finer-grained queries, which are then used to retrieve relevant information from all fields in the patient's EMRs.

We employ the BM25~\cite{robertson2009probabilistic} algorithm to compute semantic similarity, ranking and filtering field contents within the same category based on similarity scores. Fields with similarity scores exceeding 0.8 are considered valid sources. For example, if chemotherapy information for patients A and B originates from Field P of Document X (with similarity scores of 0.9 and 0.85, respectively), and for patient C from Field O of Document Y (with a similarity score of 0.95), while also appearing in Field N of Document Y (with a similarity score of 0.75), only X-P and Y-O are retained as valid sources during selection. Here, X-P appears as a source in 2/3 of cases (covering patients A and B), and Y-O appears in 1/3 of cases (covering only patient C), assigning them priorities of 2/3 and 1/3, respectively. During new patient data processing, the system first extracts content from the highest-priority field. If the field is missing, it sequentially falls back to the next most relevant field.

Ultimately, this strategy leverages semantic segmentation, similarity-based retrieval, and relevance-based filtering to refine input text, ensuring that the model generates high-quality discharge summaries that better meet clinical needs within the constraints of limited scope.

\subsection{Logic-Guided Prompt Engineering}
To suppress hallucinations caused by free-text generation while accommodating the specific needs of different medical departments, we establish explicit generation rules and constraints for various DS content types. The generation logic is categorized into five types, with corresponding optimizations applied to each:

\textbf{Extraction}: Extracts deterministic information (e.g., name, hospitalization number) for data accuracy.

\textbf{Summarization}: Summarizes key information from multiple documents (e.g., medical history) or a concise overview.

\textbf{Judgment}: Evaluates input based on clinical standards (e.g., abnormal test results) and outputs compliant conclusions.

\textbf{Inference}: Integrates data points to infer disease progression or treatment outcomes (e.g., discharge time).

\textbf{Knowledge}: Uses clinical knowledge bases to generate advisory information (e.g., follow-up departments, precautions).

To implement logic-driven DS generation, we first collaborate with medical experts to define natural language generation rules for each DS field. We then employ GPT-4o~\cite{hurst2024gpt} with a three-stage intelligent processing mechanism for optimization:

\textbf{Task Parsing}: Automatically matches generation rules with 1-4 logical structures based on predefined logic types.

\textbf{Rule Matching}: Assigns detailed generation rules to each logical structure.

\textbf{Logic Orchestration}: Integrates and generates structured, coherent, and logically sound prompt composite instructions.

Through the three-stage optimization of task parsing, rule matching, and logic orchestration, the system generates field-specific logical combination templates that comply with medical standards and maintain a clear logical flow. This enables an automated transformation from business directives to precise prompts. 
Additionally, physicians can modify the results during the rule-matching stage to meet personalized requirements. For example, if a physician wishes to include intraocular pressure test results in the DS, they can adjust the rule matching output accordingly, further optimizing the final generated content.

\subsection{Attribution-Based Comparison}
In the medical domain, the generation of discharge summaries requires clear content attribution for auditing and verification. To this end, we propose an attribution-based review method that establishes explicit correspondence between generated content and original medical records, ensuring accuracy and reliability.

Specifically, we first perform sentence-level segmentation on both the generated DS and the associated original medical records. Then, we leverage the GPT-4o model to process each generated sentence and determine its supporting sentence(s) within the original medical records. To ensure precise attribution, each sentence in the original records is assigned a unique identifier, and GPT-4o is instructed to return only the corresponding identifiers of supporting sentences.

On the user interface, when a user clicks on a sentence in the generated DS, the system highlights the corresponding original medical record sentences with the same identifier, facilitating easy comparison and verification.

\begin{table}[t]
  \centering
  \resizebox{0.48\textwidth}{!}{
  \begin{tabular}{cccc}
    \hline
    \textbf{Method} & \textbf{ROUGE-L} & \textbf{LLM-as-a-Judge} & \textbf{Human}\\
    \hline
    GPT-4o with COT      & 24.01     & 24.68         & 31.41              \\
    GPT-4o with LCDS     & 40.24      & 54.81         & 52.57              \\
    \rowcolor{blue!8}EMR\-LLM with LCDS     & \textbf{77.60}      & \textbf{75.26}          & \textbf{79.45}              \\\hline
  \end{tabular}
}
  \caption{Performance comparison of different methods, including GPT-4o with COT, GPT-4o with LCDS, and EMRLLM with LCDS. The results are evaluated using ROUGE-L, LLM-as-a-Judge, and human evaluation. The best results in each column are highlighted in bold.}
  \label{tab:evaluation}
\end{table}





\section{Evaluation}

In this section, we validate the effectiveness of LCDS through a combination of automatic and human evaluation. The experimental results are presented in Table \ref{tab:evaluation}.

\textbf{Dataset}: We collect 150 EMRs, selecting 10 from each of 15 departments.

\textbf{Baseline Methods}: To evaluate the effectiveness of LCDS, we compare it with the following three baseline methods:
1) GPT-4o with COT (Chain of Thought~\cite{wei2022chain}): Using GPT-4o for EMR-based text generation, incorporating the COT reasoning method to enhance logical consistency.
2) GPT-4o with LCDS: Using GPT-4o within the LCDS framework to optimize its performance and enhance its applicability in the medical domain.
3) EMRLLM with LCDS: Using EMRLLM within the LCDS framework to optimize DS generation and enhance output precision.

\textbf{Evaluation Metrics}: We employ both automatic and human evaluation metrics.
\textbf{Automatic Evaluation}:  
ROUGE-L~\cite{lin2004rouge} measures the longest common subsequence overlap between the generated DS and the reference DS, providing an indication of lexical similarity.  
LLM-as-a-Judge~\cite{gu2024survey} employs DeepSeek-R1~\cite{guo2025deepseek} to assess the generated text along four dimensions, including accuracy, completeness, standardization, and practicality, with a combined total score of 100 points. The evaluation criteria are detailed in Appendix~\ref{sec:appendixB}.
\textbf{Human Evaluation}:  
Medical experts assign an overall score to the generated text based on the same four dimensions, with the total score ranging from 0 to 100. Detailed evaluation guidelines are provided in Appendix~\ref{sec:appendixC}.

\textbf{Evaluation Results}: The results demonstrate that GPT-4o with LCDS outperforms GPT-4o with COT  across all metrics, indicating that the LCDS framework contributes to improved generation quality. Furthermore, EMRLLM with LCDS achieves superior performance compared to GPT-4o with LCDS, suggesting that task-specific fine-tuning on medical datasets significantly enhances generation quality.

\section{Conclusion}
We present \textbf{LCDS}, a logic-controlled discharge summary generation system that integrates precise content localization, logic-guided generation, and attribution-based expert review. By accurately extracting relevant source content, LCDS effectively reduces irrelevant information, thereby improving the quality and coherence of generated summaries. Through medical domain-specific logical constraints, LCDS significantly mitigates hallucinations and adapts to varied requirements across different clinical departments. Additionally, LCDS supports content traceability, enabling efficient expert validation, feedback, and iterative improvement of large language models in clinical practice. Our experiments on real-world clinical data demonstrate that LCDS consistently outperforms existing methods, highlighting its potential for reliable and trustworthy clinical deployment.

\section*{Limitations}
Despite the remarkable progress achieved in discharge summary generation, our study still has several limitations. First, our approach primarily relies on a specific dataset for training and evaluation, which may limit the model's generalization ability and result in degraded performance when applied to different healthcare settings or other types of electronic medical records. Second, due to the highly specialized and complex nature of medical texts, the model may generate inaccurate or ambiguous content, affecting its applicability in clinical practice. Finally, although we employ both automated and manual evaluation methods, a more comprehensive assessment of the generated text's quality and usability remains necessary. Future work could incorporate additional expert reviews or real-world clinical testing to further refine the evaluation process.

\section*{Ethics Statement}
This study strictly adheres to ethical guidelines, ensuring that all data usage complies with relevant privacy protection and data security regulations. The datasets employed have been anonymized to prevent the exposure of sensitive patient information. Additionally, we acknowledge the potential risks associated with generative models in automated medical text generation, including the possibility of producing inaccurate or misleading content. Therefore, we emphasize that the model should be used solely as an assistive tool and that all generated outputs must be rigorously reviewed and validated by medical professionals.

\section*{Acknowledgments}
We sincerely thank the anonymous reviewers for their valuable comments and suggestions. We also appreciate the support from Ruijin Hospital, Shanghai Jiaotong University School of Medicine, for this work.

\nocite{chuang2025selfcite,cohen2024contextcite,cao2024verifiable,asai2023self,hennigen2023towards,fierro2024learning,ye2023effective,zhang2024verifiable,slobodkin2024attribute}
\bibliography{custom}

\clearpage

\appendix
\section{Details on Document Types}
\label{sec:appendixA}

Our system encompasses eight types of EMR documents, including medical records, nursing records, examinations, laboratory tests, medical orders, pathology reports, diagnoses, and vital sign records. The specific content of each document type is detailed in Table~\ref{medical-info}, with representative examples available in our public repository.

To ensure consistent data representation and enable effective cross-source integration, all documents are transformed into a standardized JSON format via predefined conversion scripts upon upload. This conversion framework is designed to be both highly generalizable and configurable: by implementing tailored scripts for specific data types, we achieve precise format mapping and data normalization. Consequently, our system exhibits strong adaptability, enabling flexible application to a wide range of EMR datasets.

\begin{table*}[ht]
  \centering
  \resizebox{0.98\textwidth}{!}{
  \begin{tabular}{cccc}
    \hline
    \textbf{No.} & \textbf{Document Name} & \textbf{Content Included} & \textbf{Structure} \\
    \hline
    1 & Medical Records & Admission records, surgery records, ward round records, etc. & Unstructured data with HTML tags \\
    2 & Nursing Records & Discharge summary, etc. & XML data \\
    3 & Examination & Examination information & Structured data \\
    4 & Laboratory Test & Laboratory test information & Structured data \\
    5 & Medical Orders & Tests, prescriptions, textual reminders, etc. & Structured data \\
    6 & Pathology Report & Pathology examination information and reports & Structured data \\
    7 & Diagnosis & Diagnoses given by doctors during hospitalization & Structured data \\
    8 & Vital Signs Records & Vital signs measurements during hospitalization & Structured data \\
    \hline
  \end{tabular}
}
  \caption{\label{medical-info} Details on Document Types}
\end{table*}

\section{Evaluation Criteria for LLM-as-a-Judge}
\label{sec:appendixB}





Below is the translated version of the evaluation prompt for LLM-as-a-Judge:

Your task is to evaluate the quality of AI-generated discharge summaries (compared to the physician-written reference version).

Scoring range: 0–100 points

Scoring dimensions:

1. Information Accuracy

   - Correctness of patient identity information (e.g., name, bed number, admission number)
   
   - Accuracy of key time points (e.g., admission/discharge times)
   
   - Accuracy of brief medical history and physical examination summary at admission
   
   - Consistency of diagnostic terms with the reference answer
   
2. Medical Completeness

   - Must include core sections: brief admission history, physical exam summary, in-hospital medical course, disease progression and treatment, discharge diagnosis, medication recommendations after discharge, patient condition at discharge
   
   - Coverage of key data: laboratory tests, imaging results, surgical details, follow-up suggestions, medication guidance, etc. (no errors allowed in numerical values and test items related to the in-hospital course)
   
3. Professional Standardization

   - Standardization of medical terminology
   
   - Clear logical structure (description of diagnosis and treatment process in chronological order)
   
   - Avoid unnecessary redundancy (e.g., full-system physical examination descriptions)
   
4. Clinical Practicality

   - Actionability of discharge instructions (e.g., specific dressing change times, pathology report follow-up points)
   
   - Completeness of risk warnings (e.g., signs of incision infection)
   
Output format:

\{

  ``score'' [overall score],
  
  ``breakdown'' \{
  
    ``Information Accuracy'' [score]/40,
    
    ``Medical Completeness'' [score]/35,
    
    ``Professional Standardization'' [score]/15,
    
    ``Clinical Practicality'' [score]/10
    
  \}
  
\}

\section{Evaluation Criteria for Human}
\label{sec:appendixC}















To ensure reliable human evaluation of discharge summaries, we developed a scoring manual with a total of 100 points. The evaluation is based on four core dimensions: accuracy, completeness, standardization, and clinical utility, with an emphasis on patient safety and clinical relevance. Each dimension is scored on a scale from 0 to its maximum value;  negative scores are not permitted, and any deductions resulting in a negative value will be recorded as zero.

\subsection{Accuracy of Core Information (30 points)}
\begin{itemize}
    \item \textbf{Patient Identification}: Name, admission ID, and bed number must be correct.Each error results in a 3-point deduction.
    \item \textbf{Time Points}: Admission and discharge dates must be accurate (minute-level precision not required).Each error results in a 3-point deduction.
    \item \textbf{Diagnostic Consistency}: The discharge diagnosis must fully align with the final clinical conclusion. Descriptors like “pending paraffin section” must be included if applicable. Contradictions (e.g., benign vs. malignant misclassification) result in a 15-point deduction; omission of key diagnostic content incurs a 10-point deduction.
    \item \textbf{Admission History and Physical Exam Summary}: Should be consistent with the initial clinical documentation. Each error results in a 3-point deduction.
\end{itemize}

\subsection{Completeness of Medical Content (30 points)}
\begin{itemize}
    \item \textbf{Treatment Process Description}: Must include the procedure name, specific date, anesthesia type, and key surgical details (e.g., “right breast Mammotome excision under general anesthesia”). Missing any critical element results in an 8-point deduction.
    \item \textbf{Key Examinations During Hospitalization}: Laboratory (e.g., CBC, liver function, hepatitis panel) and imaging reports (e.g., ultrasound, chest X-ray) should be fully documented. Missing a category of essential results incurs a 5-point deduction.
    \item \textbf{Post-Discharge Instructions}: Should clearly specify pathology report follow-up timing (e.g., “10 working days”), wound care details (frequency, location, contraindications), medications, signs of complications (e.g., infection), and follow-up plans. Missing any important item leads to a 6-point deduction.
    \item \textbf{Discharge Condition}: Should be consistent with the physician’s final record; a discrepancy will result in a 5-point deduction.
\end{itemize}

\subsection{Professional Standardization (25 points)}
\begin{itemize}
    \item \textbf{Terminology}: Use standardized clinical terms (e.g., “US-BI-RADS category 3”). Each error or improper abbreviation results in a 3-point deduction.
    \item \textbf{Logical Structure}: Clinical descriptions should follow chronological order with coherent logic. Disordered descriptions result in an 8-point deduction.
    \item \textbf{Content Focus}: Irrelevant details (e.g., normal neurological exams in healthy patients) should be avoided. Redundant information results in a 5-point deduction per instance.
\end{itemize}

\subsection{Clinical Utility (15 points)}
\begin{itemize}
    \item \textbf{Actionable Recommendations}: Instructions must be specific (e.g., “change dressing on day 3 after surgery” rather than “change dressing regularly”). Vague advice results in a 5-point deduction.
    \item \textbf{Risk Mitigation}: Key complications (e.g., redness, discharge, fever) and pathology report tracking must be addressed. Missing these incurs an 8-point deduction.
    \item \textbf{Individualized Follow-up}: Abnormal findings (e.g., hepatitis B positive) should include tailored follow-up suggestions. Up to ±2 points may be adjusted based on appropriateness.
\end{itemize}

\end{document}